\documentclass[conference]{IEEEtran}
\IEEEoverridecommandlockouts
\usepackage{cite}

\usepackage{hyperref}
\usepackage{subcaption}
\usepackage{graphicx}
\usepackage{lipsum}
\usepackage{amsmath,amssymb,amsfonts}
\usepackage{graphicx}
\usepackage{textcomp}
\usepackage{xcolor}
\usepackage{algorithm}

\usepackage{multirow}
\usepackage{algorithm,algpseudocode}
\begin{document}

\title{Character Time-series Matching For Robust License Plate Recognition\\
}
\author{
\IEEEauthorblockN{Quang Huy Che\textsuperscript{1, 2}, Tung Do Thanh\textsuperscript{1, 2}, Cuong Truong Van\textsuperscript{1, 2}}

\IEEEauthorblockA{\textsuperscript{1}University of Information Technology, Ho Chi Minh City, Vietnam}
\IEEEauthorblockA{\textsuperscript{2}Vietnam National University, Ho Chi Minh City, Vietnam}

Email: \{18520819, 19522491\}@gm.uit.edu.vn, cuongtv@uit.edu.vn
}

\maketitle

\begin{abstract}
Automatic License Plate Recognition (ALPR) is becoming a popular study area and is applied in many fields such as transportation or smart city. However, there are still several limitations when applying many current methods to practical problems due to the variation in real-world situations such as light changes, unclear License Plate (LP) characters, and image quality. Almost recent ALPR algorithms process on a single frame, which reduces accuracy in case of worse image quality. This paper presents methods to improve license plate recognition accuracy by tracking the license plate in multiple frames. First, the Adaptive License Plate Rotation algorithm is applied to correctly align the detected license plate. Second, we propose a method called Character Time-series Matching to recognize license plate characters from many consequence frames. The proposed method archives high performance in the UFPR-ALPR dataset which is \boldmath$96.7\%$ accuracy in real-time on RTX A5000 GPU card. We also deploy the algorithm for the Vietnamese ALPR system. The accuracy for license plate detection and character recognition are 0.881 and 0.979 $mAP^{test}$@.5 respectively. The source code is available at \url{https://github.com/chequanghuy/Character-Time-series-Matching.git}

\end{abstract}


\section{Introduction} 

An ALPR system has targeted traffic management in the modern world since its first introduction in \cite{1990}. Many approaches have been introduced to make a system possible to predict LP location and identify the information of each LP most accurately. This increases the importance of an ALPR system in various applications such as automatic traffic toll systems, traffic management systems, automatic handling violations systems, etc.  

A general ALPR system has two main steps, which are License Plate detection and License Plate recognition. For the License Plate detection step, all LPs were extracted from the image. In this step, most recent works prefer to apply CNN-based object detection methods such as R-CNN \cite{rcnn}, Mask R-CNN \cite{maskrcnn}, etc., to locate the region of vehicles that have detected license plates. In the next step, License Plate recognition, characters are recognized by using a Character Segmentation network with a classification head like CRNET \cite{crnet} or Optical Character Recognition \cite{ocr}. Although the result of each step achieves state-of-the-art accuracy, however, the overall accuracy of a whole system is still limited. 

 The authors in \cite{layout} proposed an ALPR system that obtained SOTA on many public datasets except for the UFPR-ALPR dataset. Due to the variety of inclement conditions (lighting, noise, camera distortion) in UFPR-ALPR. We noticed that besides the high performance of their previous stage, there are still some missing characters that devastate the final prediction. Therefore, it influences the character recognition result. The temporal redundancy method, which has been applied in papers \cite{ufpr,layout,temporal} has not yet taken advantage of all the license plate information. To resolve this problem, we propose a Character Time-series Matching method based on the Hungarian method and an Adaptive License Plate Rotation method to improve the ALPR system performance in various conditions. 

It is also worth noting that we decide to evaluate the system only on the UFPR-ALPR dataset for many reasons. First, ALPR systems typically analyze videos or image sequences, and it is essential to use appropriate datasets to assess the accuracy of the whole system, such as the UFPR-ALPR dataset, which includes multiple video tracks. Second, this dataset has a lot of complex cases similar to a natural environment. Although, the paper \cite{layout} achieved the average end-to-end recognition rate of $96.9\%$ across eight public datasets. However, as regards the UFPR-ALPR dataset, it was the lowest accuracy, at just about $90\%$ \cite{layout}.

In summary, our contributions can be given as follows: 

\begin{itemize} 

\item A complete ALPR system based on the YOLOv5 model for object detection tasks. 

We then propose two methods called Character Time-series Matching and Adaptive License Plate Rotation which increase the accuracy of Character Recognition under challenging conditions.

\item Regarding the result assessment, we evaluate our system on the UFPR-ALPR public dataset and our private dataset, which is gathered from multiple IP cameras on the streets of Vietnam. The proposed system archives excellent results on the UFPR-ALPR dataset and can process in real-time, both mid-end and high-end GPUs. In addition, we also implement the system on the Jetson Nano board and Jetson TX2 board for system performance evaluation. 
\end{itemize}

\section{Related work}
This section introduces some state-of-the-art approaches to the ALPR system. Most research includes two main tasks: Vehicle-License Plate Detection and License Plate Recognition.

\subsection{Vehicle and License Plate Detection}

License plates can be extracted by using an object detection model. Two popular methods are the one-stage and two-stage object detector model. As for two-stage object detection, Ross Girshick et al proposed R-CNN \cite{rcnn} model that uses the selective search algorithm to extract the region proposals in the first stage. Then a CNN model will be used to classify the objects in the second stage. Similarly, the Fast R-CNN model \cite{fastrcnn} based on R-CNN has a great soar in both speed and accuracy resulting from applying Region of Interest on feature maps. 

Another model named Faster R-CNN \cite{fasterrcnn} uses the Region Proposal Network to reduce the computational time considerably, which is ten times and fifty times compared to Fast R-CNN and R-CNN. However, these models are hard to process in real-time when compared to one-stage object detection; the YOLO \cite{YOLOv5} and SSD \cite{ssd} get much better both speed and accuracy. 

In the paper \cite{ufpr, layout}, the vehicle is first detected, then the license plate is cropped for the next processing step by using another object detection network. It can be seen that the authors need to use two CNNs models for this task which leads to increased resource usage and processing time. In \cite{brazilian}, instead of using the vehicle annotation, they expand the LP's annotation and call that extended area the Frontal View. The model learns the frontal views instead of using vehicle annotation and then takes those areas to detect LPs. Paper \cite{lite} used a multi-tasking CNN (MTCNN) model, which combines 3 CNNs, P-net, R-net, and O-net, to improve the model performance. As a result, this method is generally slower than single-stage detector models. 

\subsection{License Plate Recognition}
Regarding license plate recognition, there are two primary trends which are segmentation-based and segmentation-free approaches. In the segmentation-based method, each detected license plate is fed to the segmentation network to locate characters such as CRNet \cite{ufpr}  and DeepLabv2 ResNet-101 \cite{deeplab}. After that, to recognize characters, an OCR or a Character Classification network is used. In the segmentation-free method, \cite{lite} locates characters by using the Convolutional Recurrent Neural network, and then the Connectionist Temporal Classification is applied to the next step. In \cite{vertex}, they employ the resample and rectified LP and the SRNET instead of Recurrent Neural networks because of growth performance in the same resource usage. While instead of using the segmentation-based method, the authors \cite{layout} propose a method for recognizing characters based on the CRNet model. Due to the integration of localization and classification into a single step, the segmentation-free method is faster than segmentation methods. In practical applications, processing on a single frame easily makes erroneous results. To resolve this problem, in papers \cite{ufpr, layout}, the authors apply the Temporal Redundancy technique to the sequential frames to get better results. 

\section{Proposed method}
\begin{figure*}[t]
\centering
\includegraphics[width=\linewidth]{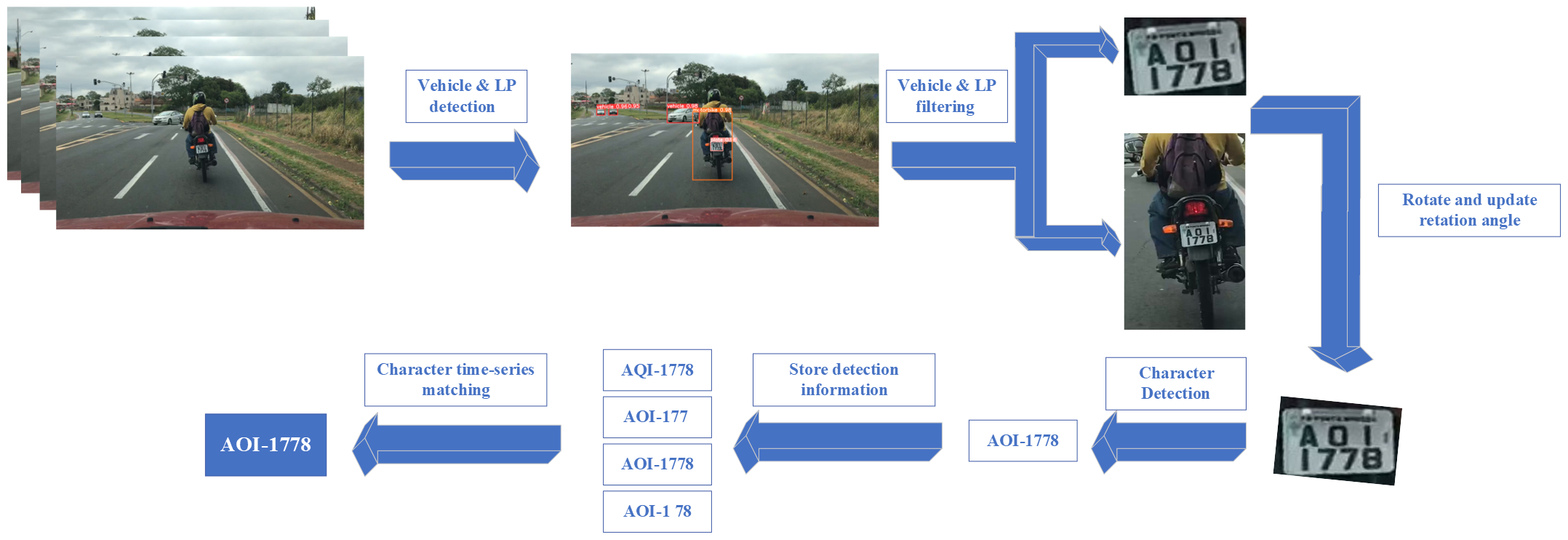}
\caption{ALPR pipeline}    
\label{ALPR pipeline}

\end{figure*}

This paper implements an ALPR system for applying our proposed method. Figure.~\ref{ALPR pipeline} illustrates our proposed system. The system includes two steps which are License Plate Detection and Character Detection. In the first step, all vehicles with match license plates are extracted. Next, all license plate characters will be recognized in the License Plate Recognition. In order to improve the recognition result, the Adaptive License Plate Rotation algorithm is applied to align the license plate image. Moreover, we also propose the Character Time-series Matching algorithm based on tracking characters in continuous frames for archiving better results. The detail of our proposed methods is discussed below sections.

\subsection{License Plate Detection}
Recent works divide this step to separate stages: vehicle detection and license plate detection \cite{ufpr, layout}. It leads to an increase marginally in both computational cost and processing time. Thanks to YOLOV5 model for detecting small objects well, we design a detection model based on YOLOv5m model with an input size of 1,280$\times$720. License plates and Vehicles are detected simultaneously in real-time. We train the model based on pretrained YOLOv5m with our custom dataset, which includes all vehicles and license plate annotations.

\subsection{Character Detection}

In the Character Detection step, firstly, all detected license plates are stored and sorted by time in a queue. After the tracked object disappears, we propose an additional post-processing algorithm that finds vehicles match for license plate and the Adaptive License Plate Rotation method to align the license plate image. Eventually, we apply the Character Time-series Matching algorithm to get the final result. Regarding the character detection model, we propose a model based on YOLOv5 with the multi-head self-attention block proposed in the paper\cite{attention}, and the input size is $128 \times 128$. However, since the characters are relatively large compared to the license plate size, we drop the detection head for the small and medium features and halve the number of filters in every layer, resulting in only slightly reducing the detection accuracy but significantly improving the speed. Table \ref{table1} shows the architectural design of our character detection model, where a $C3$ module is three consecutive Convolution layers with a Bottleneck block at the end, and $C3TR$ is the $C3$ module that replaced the Bottleneck block with a multi-head self-attention layer.

\subsubsection{The Adaptive License Plate Rotation method}



\begin{figure}[htp]
  \centering
    \includegraphics[width=0.58\textwidth]{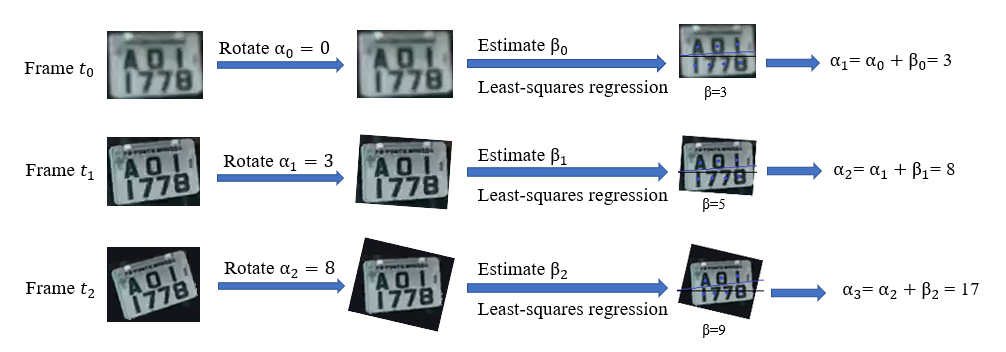}
  \caption{Visual of The Adaptive License Plate Rotation \\
    (a) \emph{Original Images} (b) \emph{Rotated images} (c) \emph{Alignment images and updated values} 
  }
  \label{rotation}
\end{figure}

In the practical environment, vehicles can move in many different directions while the camera view is fixed in one spot. Fig.~\ref{rotation} - (a) shows some samples in this case. Therefore, the license plate image may be tilted in a particular direction depending on the vehicle's direction. It makes a considerable reduction in the accuracy of the next step. To handle this issue, we need to adjust the angle of the license plate correctly. Assume there is a line $d$ that divides the license plate into two equal parts horizontally, which form by the equation $y=ax+b$, $a$ is the slope, $\alpha$ is the angle formed by the horizontal axis, and the line $d$, so that $\alpha=\arctan{a}$. Fig.~\ref{rotation} illustrates the Adaptive License Plate Rotation method, in which (a) are original images and (b) are rotated images. 

In the first frame $t_0$, we initialize $\alpha_0=0$ and estimate the $\beta_0 = \arctan({a_0})$, where $a_0$ is calculated following equation \eqref{linear}. In the following frame, the LP image is rotated $\alpha_1$-degree angle, where $\alpha_1$ is updated based on equation \eqref{rotate}, then the character prediction model predicts the bounding boxes from the rotated LP image at time $t_1$ and estimates the $\beta_1$ value. This process repeats until the final LP frame. 

\begin{equation}
\label{linear}
    a_t =\frac{\sum_{i=1}^{n_t} x_i y_i - n \bar{x} \bar{y}}{\sum_{i=1}^{n_t} x^2_i-n(\bar{x})^2}
\end{equation}
\begin{itemize}[where:]
    \item $n_t$: number of character bounding boxes from LP image at time $t$.
    \item $x_i, y_i$: $x$ center and $y$ center of the character bounding box $i$.
    \item $a_t$: the slope coefficient of the line $d$ at time $t$.
\end{itemize}

\begin{equation}
\label{rotate}
\alpha_t=\alpha_{t-1}+\beta_{t-1}
\end{equation}
\begin{itemize}[where:]
    \item $\alpha_{t}$: the estimated angle between LP image and the horizontal axis at time $t-1$.
    \item $\alpha_{t-1}$: the estimated angle between LP image and the horizontal axis at time $t-1$.
    \item $\beta_{t-1}$: the estimated angle between rotated LP image and the horizontal axis at time $t-1$.
\end{itemize}

\begin{table}[h!]
  \begin{center}
    \centering
    \caption{Overview of the proposed model.}
    \label{tab:table1}
    \begin{tabular}{|p{8mm}|p{12mm}|p{13mm}|p{18mm}|p{16mm}|}
    \hline
      \textbf{Module index} & \textbf{Connected from} & \textbf{Module}& \textbf{Input size}& \textbf{Output size}\\
      \hline
      0 &   & Conv & $3\times128\times128$ & $16\times64\times64$\\
      1 & 0 & Conv & $16\times64\times64$ &$32\times32\times32$ \\
      2 & 1 & C3 & $32\times32\times32$ & $32\times32\times32$ \\
      3 & 2 & Conv &$32\times32\times32$ &$64\times16\times16$ \\
      4 & 3 & C3 &$64\times16\times16$ &$64\times16\times16$ \\
      5 & 4 & Conv &$64\times16\times16$ &$128\times8\times8$ \\
      6 & 5 & C3 &$128\times8\times8$ &$128\times8\times8$ \\
      7 & 6 & Conv &$128\times8\times8$ &$256\times4\times4$ \\
      8 & 7 & SPP &$256\times4\times4$ &$256\times4\times4$ \\
      9 & 8 & C3TR &$256\times4\times4$ &$256\times4\times4$ \\
      10 & 9 & Conv &$256\times4\times4$ &$128\times4\times4$ \\
      11 & 10 & Upsample &$128\times4\times4$ &$128\times8\times8$ \\
      12 & [6,11] & Concatenate & $[128\times8\times8]$\newline$[128\times8\times8]$ &$256\times8\times8$ \\
      13 & 12 & C3 &$256\times8\times8$ &$128\times8\times8$ \\
      14 & 13 & Conv &$128\times8\times8$ &$64\times8\times8$ \\
      15 & 14 & Upsample &$64\times8\times8$ &$64\times16\times16$ \\
      16 & [4,15] & Concatenate & $[64\times16\times16]$\newline$[64\times16\times16]$ &$128\times16\times16$ \\
      17 & 16 & C3 &$128\times16\times16$ &$64\times16\times16$ \\
      18 & 17 & Detect &$64\times16\times16$ & \\
      \hline
    \end{tabular}
    \label{table1}
  \end{center}
\end{table}
\subsubsection{Character Time-series Matching}\label{SCM}

\begin{figure}[htp]
  \centering
    \includegraphics[width=0.5\textwidth]{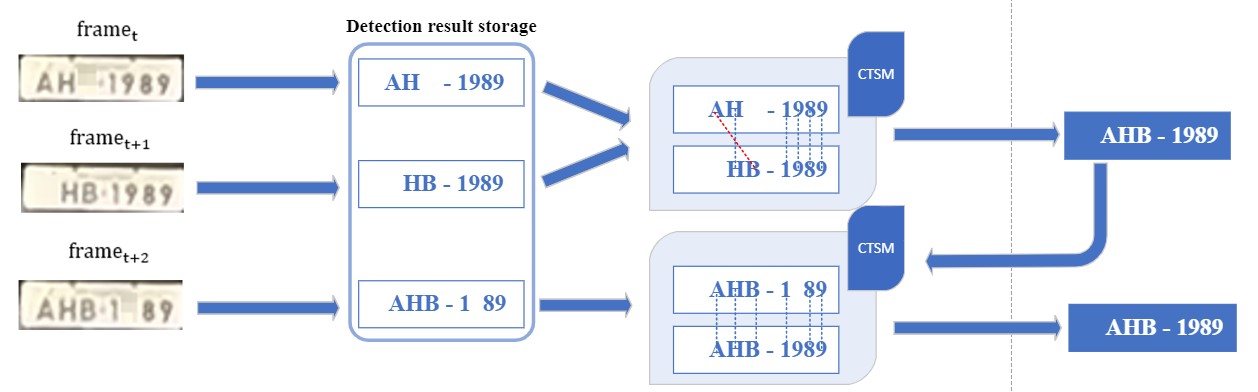}
  \caption{Character Time-series Matching (\textbf{CTM})}
  \label{ctsm}
\end{figure}

In our experiments, processing on a single frame achieves poor results in the case of image quality. Therefore, in this paper, the Character Time-series Matching (CTM) algorithm is proposed to take advantage of the spatial information obtained. CTM combines two bounding boxes of the same character in the same LP from 2 consecutive frames based on the Euclidean distance between the center of the two bounding boxes, depicted in Fig.~\ref{ctsm}.  
In CTM, if two character bounding boxes are joined together, the Euclidean distance between its center boxes  must be less than the threshold $\epsilon$.


In detail, we describe our problem as a linear assignment problem, with the objects assigned to each other being the character bounding boxes from $I^{a}_{t}$ and $I^{a}_{t+1}$, where $I^{a}_{t}$ and $I^{a}_{t+1}$ are the consecutive images of the LP $a$ at time $t$ and $(t+1)$, respectively. The cost for the linear assignment problem is the Euclidean distance between the two bounding boxes. We solve the problem by using the modified Hungarian algorithm \cite{hungarian}. 

\begin{figure}[htp]
  \centering
    \includegraphics[width=0.5\textwidth]{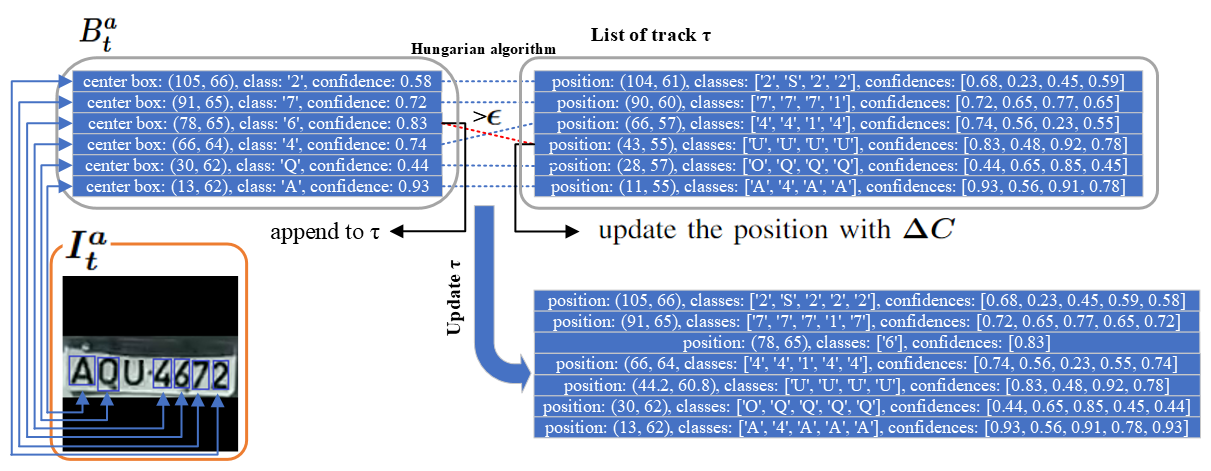}
  \caption{The update process of Character Time-series Matching}
  \label{updatebox}
\end{figure}

At first, a set of track is initialized, $\tau =\{\}$ based on sort algorithm \cite{sort}, with each track $t \in \tau$ containing the current position (the center of the bounding box), the list $cls$ contains the class of the bounding boxes that have been matched with track $t$ in the previous frame and the list $conf$ stores the confidence corresponding to each class in the $cls$.
We consider three conditions:

\begin{itemize}[]
    \item If a bounding box $b_j$ from $B^{a}_{t}$ matches with a track $t_i$ from $\tau$, the result of $t_i$ is updated with the position of $t_i$ assigned to the position of $b_j$ ($t_{i_{position}}=b_{j_{position}}$), the character class and confidence of $b_j$ is appended to the corresponding list in  $t_i$;  where $B^{a}_{t}$ is the bounding boxes from the Character Detection stage of $I^{a}_{t}$.
    
    \item If a bounding box $b_k$ from $B^{a}_{t}$ does not match any track from $\tau$, then $b_k$  is appended to set $\tau$. 

    \item If a track $t_m$ of $\tau$ does not match with any bounding box of $B^{a}_{t}$, $t_m$ is added a $\Delta C$ value; where $\Delta C$ is the mean distances between the position of all possible track and bounding boxes that can be matched. The idea of updating $\Delta C$ is that we want to shift the ``unmatchable" tracks in the same direction as the ``matchable" tracks.
   
\end{itemize}
Fig.~\ref{updatebox} shows information about how the Character Time-series Matching update parameters. 

\begin{figure}[htp]
  \centering
    \includegraphics[scale=0.6]{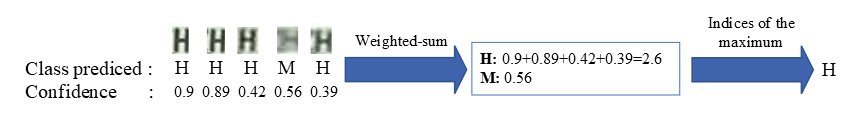}
  \caption{The character result based on the weighted-sum confidence of each class}
  \label{weightsum}
\end{figure}

We perform post-processing to get the character detection result of each track in the set $\tau$  by calculating the weighted-sum confidence of each class and then select the character class with the most confidence value, as shown in Fig.~\ref{weightsum}. As shown in Fig.~\ref{weightsum}, character H is selected, and character M is removed because its weighted-sum value is lower than character H. Equation \ref{weighted} shows the detail of this step.

\begin{equation}
\label{weighted}
\resizebox{1\hsize}{!}{$class_i=\underset{c\in class-indices}{\arg\max}\{K^c_i|K^c_i={\sum_{\substack{j}}class_i^j confidence_i^j}\}$}
\end{equation}

\begin{itemize}[where:]
    \item $class-indices$: the list of classes' indices
    \item $class_i$: class of $track_i$
    \item $K_i^c$: weighted-sum confidence of class c in $track_i$
    \item $class_i^j$: is set to 1 if the class of the $j^{th}$ box detection in track i is $c$
    \item $confidence_i^j$: confidence of the $j^{th}$ box detection in $track_i$
\end{itemize}

\section{experimental results}
This section evaluates the whole pipeline on the UFPR-ALPR and Vietnamese streets to ensure that the suggested ALPR system is effective. All tests were carried out using Pytorch framework on NVIDIA GeForce RTX A5000, RAM 32GB, Intel(R) Core(TM) i9-10900X.

\begin{figure}[htp]
  \centering
    \begin{subfigure}{0.45\columnwidth}
        \includegraphics[width=\textwidth]{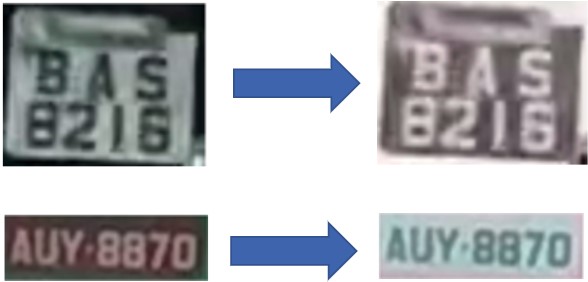}
        \caption{Negative images} 
    \end{subfigure} 
    \begin{subfigure}{0.45\columnwidth}
        \includegraphics[width=\textwidth]{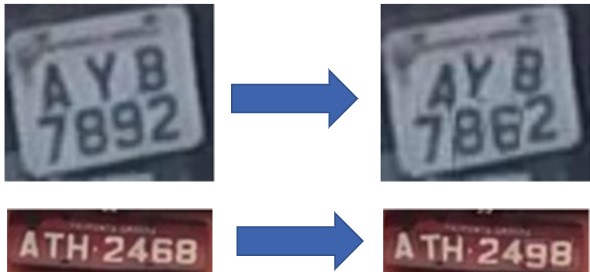}
        \caption{Flip Characters} 
    \end{subfigure} 
    \caption{Illustrated image of data augmentation methods}
    \label{flipchar}
\end{figure}


\subsection{Evaluation on the UFPR-ALPR dataset}

\subsubsection{Vehicle and License Plate Detection}
Although we used only one model for vehicle and LP detection, we achieved competitive accuracy compared to the \cite{layout} using two models, including Vehicle Detection and License Plate Detection. The model accuracy results obtained after applying the filter are shown in Table \ref{table2}.
\begin{table}[h!]
  \begin{center}
    \centering
    \caption{Accuracy and inference time in Vehicle and License Plate Detection stage.}
    \label{tab:table2}
    
    \begin{tabular}{|p{11mm} | p{11mm} | p{8mm} | p{12mm} | p{11mm}|}
    \hline
      \textbf{Size} & \textbf{Precision} & \textbf{Recall}& \textbf{mAP\newline test@.5}&  \textbf{Inference time}\\
      \hline
      1280$\times$720 & 0.968  & 0.959 & 0.968 & 12.4ms\\
      \hline
    \end{tabular}
    \label{table2}
  \end{center}
\end{table}

\subsubsection{Character Detection}
Due to data restrictions for characters, we proceed to gather and categorize additional data, including over 900 images of Brazilian LPs\footnote{The images were downloaded from \url{www.platesmania.com}}. Although our suggested model only comprises $1213854$ parameters and $0.15$ GFLOP with an input image size of $128\time128$, it has a good level of accuracy on the UFPR-ALPR dataset's test set.

The proposed model has a much lower computational cost than the previous character recognition model \cite{layout} \cite{ufpr} so that it can be efficiently run on low-configuration machines and embedded computers.

We consider the characters `1' and `I', and the characters `0' and `O' as single classes in the assessments performed in the UFPR-ALPR dataset since they are similar depending on the format of the LP that we adjust accordingly. The Brazilian LP format is AAA-NNNN, where A is an alphabetic character and N is a numeric character.

We use data augmentation methods such as rotation, mosaic, and mix-up and also apply data augmentation by converting negative images and randomly flipping characters directly on number plate images for the training process. The suggested characters to be rotated are listed in Table \ref{table3}. The data augmentation technique automatically transforms the class of character `6' to character `9' and the class of character `9' to character `6' when flip character `6' or character `9' in both directions. Some instances of the proposed data augmentation approach are shown in Fig.~\ref{flipchar}.
\begin{table}[h!]
  \begin{center}
    \centering
    \caption{Statistical table of characters that can be flipped in each direction.}
    \begin{tabular}{|c | c |}
    \hline
      \textbf{Flip Direction} & \textbf{Characters} \\
      \hline
      Vertical & 0 (O), 1 (I), 3, 8, B, C, D, E, H, K, X   \\
      Horizontal &  0 (O), 1 (I), 8, A, H, M, T, U, V, W, X, Y  \\
      Both &  0 (O), 1 (I), 6, 8, 9, H, N, S, X, Z   \\
     \hline
    \end{tabular}
    \label{table3}
  \end{center}
\end{table}

Table \ref{table4} is the result of the proposed model on the test set of the UFPR-ALPR dataset, with a confidence threshold of 0.1 and an IOU threshold of 0.1.

\begin{table}[h!]
  \begin{center}
    \centering
    \caption{Accuracy and inference time in Character Detection stage.}
    \begin{tabular}{|p{11mm} | p{11mm} | p{8mm} | p{12mm} | p{12mm}| p{11mm}|}
    \hline
      \textbf{Size}  & \textbf{Precision} & \textbf{Recall}& \textbf{mAP\newline test@.5}& \textbf{Inference time}\\
      \hline
        128$\times$128  & 0.95 & 0.964 & 0.984 & 6.1ms\\
      \hline
    \end{tabular}
    \label{table4}
  \end{center}
\end{table}
\subsubsection{ALPR system (end to end)}
\begin{figure}[ht] 
  \begin{subfigure}[b]{0.33\linewidth}
    \centering
    \captionsetup{justification=centering}
    \includegraphics[width=.62\linewidth]{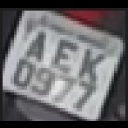} 
    \caption{Track 105\\
    \cite{ufpr}: -\\
    \cite{layout}: -\\
    {\fontsize{8}{8}\selectfont CTM:A\textcolor{red}{F}K-0977}\\
    {\fontsize{8}{8}\selectfont AR+CTM: AEK-0977}}
    \label{compare:a} 
    \vspace{2ex}
  \end{subfigure}
  \begin{subfigure}[b]{0.33\linewidth}
    \centering
    \captionsetup{justification=centering}
    \includegraphics[width=.62\linewidth]{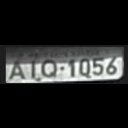} 
    \caption{Track 111\\
    \cite{ufpr}: A\textcolor{red}{UC}1056\\
    \cite{layout}: A\textcolor{red}{T}Q1056\\
    {\fontsize{8}{8}\selectfont CTM: AIQ1056}\\
    {\fontsize{8}{8}\selectfont AR+CTM: AIQ1056}}
    \label{compare:b} 
    \vspace{2ex}
  \end{subfigure} 
  \begin{subfigure}[b]{0.33\linewidth}
    \centering
    \captionsetup{justification=centering}
    \includegraphics[width=.62\linewidth]{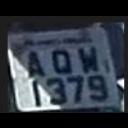} 
    \caption{Track 127\\
    \cite{ufpr}: {\fontsize{8}{8}\selectfont \textcolor{red}{-}OW\textcolor{red}{7}379}\\
    \cite{layout}: {\fontsize{8}{8}\selectfont A\textcolor{red}{O}W1379}\\
    {\fontsize{8}{8}\selectfont CTM: AQW1379}\\
    {\fontsize{8}{8}\selectfont AR+CTM: AQW1379}}
    \label{compare:b} 
    \vspace{2ex}
  \end{subfigure} 
  \begin{subfigure}[b]{0.33\linewidth}
    \centering
    \captionsetup{justification=centering}
    \includegraphics[width=0.62\linewidth]{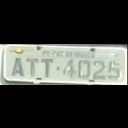} 
    \caption{Track 131 \\
    \cite{ufpr}: {\fontsize{8}{8}\selectfont AT\textcolor{red}{U}4025}\\
    \cite{layout}: {\fontsize{8}{8}\selectfont ATT402\textcolor{red}{6}}\\
    {\fontsize{8}{8}\selectfont CTM: ATT4025}\\
    {\fontsize{8}{8}\selectfont AR+CTM: ATT4025}}
    \label{compare:c} 
  \end{subfigure}
  \begin{subfigure}[b]{0.33\linewidth}
    \centering
    \captionsetup{justification=centering}
    \includegraphics[width=0.62\linewidth]{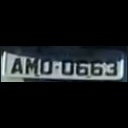} 
    \caption{Track 136\\
    \cite{ufpr}: {\fontsize{8}{8}\selectfont AMO0663}\\
    \cite{layout}: {\fontsize{8}{8}\selectfont AMO0663}\\
    {\fontsize{8}{8}\selectfont CTM: AMO0663\textcolor{red}{3}}\\
    {\fontsize{8}{8}\selectfont AR+CTM: AMO0663\textcolor{red}{3}}}
    \label{compare:d} 
  \end{subfigure} 
  \begin{subfigure}[b]{0.33\linewidth}
    \centering
    \captionsetup{justification=centering}
    \includegraphics[width=0.62\linewidth]{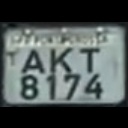} 
    \caption{Track 148\\
    \cite{ufpr}: -\\
    \cite{layout}: -\\
    {\fontsize{8}{8}\selectfont CTM: AKT8\textcolor{red}{7}74}\\
    {\fontsize{8}{8}\selectfont AR+CTM: AKT8174}}
    \label{compare:d} 
  \end{subfigure} 
  \caption{Some results were obtained on \cite{ufpr}, \cite{layout}, and the proposed method.}
  \label{compare} 
\end{figure}

Table \ref{table5} shows the result of our proposed system compared to the others on the UFPR-ALPR dataset. In the case of only using the CTM method, the figure for accuracy is 91.67 compared to nearly 90 and 64.9 in paper \cite{layout} and \cite{ufpr}, respectively. Another interesting point is a considerable increase in case integration of both AR and CTR methods, with 96.7 reaching the highest accuracy. The results are at around 47 fps in our system. Fig.~\ref{compare} shows some results when applying the proposed algorithm to the UFPR-ALPR dataset.


\begin{table}[h!]

\centering
\caption{The accuracy of ALPR system.}
\scalebox{0.85}{%

\begin{tabular}[t]{lcccc}
\hline
& \textbf{\cite{ufpr}} & \textbf{\cite{layout}}&\textbf{CTM} & \textbf{AR $^{\mathrm{a}}$ + CTM}\\
\hline
Number of used models & 4& 3 & \textbf{2} & \textbf{2}\\
Accuracy&64.9&90.0 ± 0.7& 91.67 & \textbf{96.7} \\
FPS&-&-& \textbf{47.6} & 47.56\\
\hline
\multicolumn{4}{l}{$^{\mathrm{a}}$ Our baseline with the Adaptive License Plate Rotation method.}
\label{table5}
\end{tabular}

}
\end{table}

\begin{figure*}[htp]
  \centering
    \includegraphics[scale=0.4]{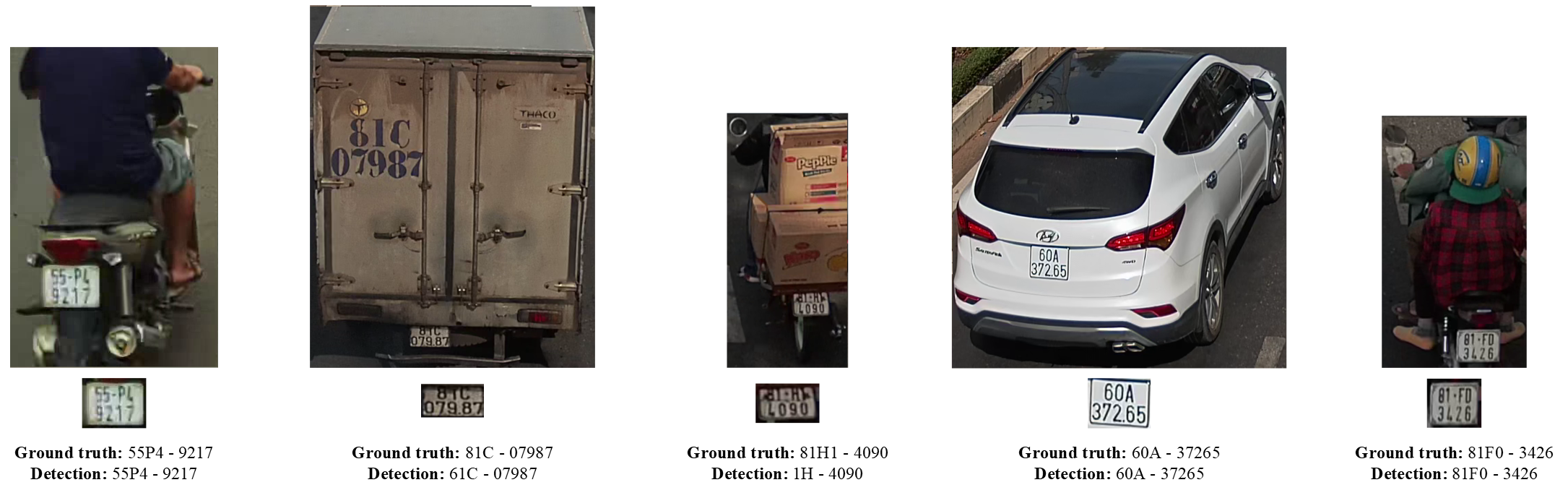}
  \caption{Some qualitative results obtained from our test video}
  \label{testcase}
\end{figure*}
\subsection{Evaluation on Vietnamese street}
In addition to developing a system to provide a powerful LP Recognition method, we collected and labeled 9,887 Vietnamese license plate images, which have two types: square and rectangular license plates and 28180 images for LP recognition on Vietnamese streets. 

We also evaluated on the Jetson Nano kit and Jetson TX2 kit, with the results obtained as shown in Table \ref{table6}. Overall, what stands out from the table is that our proposed method can be implemented well on embedded computers, and the execution time is more noticeable when converted to 16-bit TensorRT floating point inference.
\begin{table}[]
\caption{Evaluation results on Vietnamese dataset.}
\scalebox{0.8}{
\begin{tabular}{|c|c|c|c|c|ccc|}

\hline
\multirow{2}{*}{Model}                                                         & \multirow{2}{*}{\begin{tabular}[c]{@{}c@{}}Inference \\ type\end{tabular}} & \multirow{2}{*}{Precision} & \multirow{2}{*}{Recall} & \multirow{2}{*}{\begin{tabular}[c]{@{}c@{}}mAP \\ test@.5\end{tabular}} & \multicolumn{3}{c|}{Inference time}                               \\ \cline{6-8} 
                                                                               &                                                                            &                            &                         &                                                                         & \multicolumn{1}{c|}{Nano}   & \multicolumn{1}{c|}{TX2}   & Server \\ \hline
\multirow{2}{*}{\begin{tabular}[c]{@{}c@{}}YoloV5s \\ 640x640\end{tabular}}    & Pytorch                                                                    & 0.876                      & 0.809                   & 0.847                                                                   & \multicolumn{1}{c|}{198ms}  & \multicolumn{1}{c|}{78ms}  & 8.3ms  \\ \cline{2-8} 
                                                                               & TensorRT                                                                   & 0.884                      & 0.789                   & 0.841                                                                   & \multicolumn{1}{c|}{71ms}   & \multicolumn{1}{c|}{33ms}  & 0.9ms  \\ \hline
\multirow{2}{*}{\begin{tabular}[c]{@{}c@{}}YoloV5m\\ 640x640\end{tabular}}     & Pytorch                                                                    & 0.875                      & 0.823                   & 0.866                                                                   & \multicolumn{1}{c|}{515ms}  & \multicolumn{1}{c|}{180ms} & 10.9ms \\ \cline{2-8} 
                                                                               & TensorRT                                                                   & 0.859                      & 0.829                   & 0.867                                                                   & \multicolumn{1}{c|}{171ms}  & \multicolumn{1}{c|}{75ms}  & 1.9ms  \\ \hline
\multirow{2}{*}{\begin{tabular}[c]{@{}c@{}}YoloV5m\\ 1280x1280\end{tabular}}   & Pytorch                                                                    & 0.877                      & 0.839                   & 0.881                                                                   & \multicolumn{1}{c|}{2162ms} & \multicolumn{1}{c|}{855ms} & 15.6ms \\ \cline{2-8} 
                                                                               & TensorRT                                                                   & 0.869                      & 0.835                   & 0.879                                                                   & \multicolumn{1}{c|}{615ms}  & \multicolumn{1}{c|}{244ms} & 5.3ms  \\ \hline
\multirow{2}{*}{\begin{tabular}[c]{@{}c@{}}Yolov5+\\ Transformer\end{tabular}} & Pytorch                                                                    & 0.986                      & 0.971                   & 0.979                                                                   & \multicolumn{1}{c|}{30ms}   & \multicolumn{1}{c|}{25ms}  & 6.1ms  \\ \cline{2-8} 
                                                                               & TensorRT                                                                   & 0.898                      & 0.893                   & 0.822                                                                   & \multicolumn{1}{c|}{11ms}   & \multicolumn{1}{c|}{9.5ms} & 0.75ms \\ \hline
\end{tabular}}
\label{table6}
\end{table}

Moreover, we test on a set of 100 videos recorded from IP cameras in real-world conditions. If the system successfully detects the license plate information and vehicle type, the case is said to be true in our test set. Our system achieves $90\%$ accuracy, meaning it could extract correctly $90$ LP information and vehicle type from $90$ vehicles; Fig.~\ref{testcase} shows some sample images and results. 



\section{Conclusion}
In conclusion, we propose the Character Time-series Matching and Adaptive Rotation algorithm, which improves the accuracy of the ALPR system. The algorithms archives high accuracy in case of low quality and small objects. In addition, the processing time of our method can perform in real-time in mid-end and high-end GPU. The experiments on UFPR-ALPR and Vietnamese streets show that our proposed method has excellent performance in various conditions. In the future, we would like to test the effectiveness of our approach on more public datasets and try to optimize CNN architectures for processing on edge computers.

{\small
\bibliographystyle{IEEEtran}
\bibliography{ref}
}

\end{document}